\documentclass[preprint,12pt]{elsarticle}

\usepackage{palatino,epsfig,latexsym}
\usepackage{natbib}
\usepackage{amssymb}
\usepackage{amsmath}
\usepackage{amsthm}
\usepackage{graphicx}
\usepackage{xcolor}
\graphicspath{{figures/}}
\usepackage{subfigure}
\usepackage[ruled,vlined,linesnumbered]{algorithm2e}
\usepackage{url}

\newcommand{\tabincell}[2]{\begin{tabular}{@{}#1@{}}#2\end{tabular}}

\begin{document}

\begin{frontmatter}



\title{Simplex Search Based Brain Storm Optimization\tnoteref{jijin}}
\tnotetext[jijin]{This work was supported by National Key R\&D Program of China (No. 2016YFD0400206), NSF of China (No. 61773119) and NSF of Guangdong Province (No. 2015A030313648).}
\author[label1]{Wei Chen}
\author[label1]{YingYing Cao}
\author[label2]{Shi Cheng}
\author[label3,label4]{Yifei Sun}
\author[label1]{Qunfeng Liu\tnoteref{cauthor}}
\address[label1]{School of Computer Science and Network Security, Dongguan University of Technology, Dongguan, 523808, China}
\address[label2]{School of Computer Science, Shaanxi Normal University, Xi'an, 710119, China}
\address[label3]{Key Laboratory of Modern Teaching Technology, Ministry of Education, Shaanxi Normal University, Xi'an, 710062, China}
\address[label4]{School of Physics and Information Technology, Shaanxi Normal University, Xi'an, 710119, China}
\tnotetext[cauthor]{Corresponding author: Qunfeng Liu (liuqf@dgut.edu.cn) and Yun Li (Yun.Li@ieee.org)}
\author[label1]{Yun Li\tnoteref{cauthor}}

\begin{abstract}
Through modeling human's brainstorming process, the brain storm optimization (BSO) algorithm has become a promising population-based evolutionary algorithm. However, BSO is pointed out that it possesses a degenerated L-curve phenomenon, i.e., it often gets near optimum quickly but needs much more cost to improve the accuracy. To overcome this question in this paper, an excellent direct search based local solver, the Nelder-Mead Simplex (NMS) method is adopted in BSO. Through combining BSO's exploration ability and NMS's exploitation ability together, a simplex search based BSO (Simplex-BSO) is developed via a better balance between global exploration and local exploitation. Simplex-BSO is shown to be able to eliminate the degenerated L-curve phenomenon on unimodal functions, and alleviate significantly this phenomenon on multimodal functions. Large number of experimental results show that Simplex-BSO is a promising algorithm for global optimization problems.
\end{abstract}


\end{frontmatter}


\section{Introduction}\label{section-1}
Numerous scientific or engineering problems can be modeled as the following optimization problem
\begin{equation}\label{question}
\min ~f(x), ~ s.t., ~x\in\Omega\subseteq \mathbb{R}^n,
\end{equation}
where $n$ is the number of controllable variables, and $f(x)$ is the objective function. When $\Omega = \mathbb{R}^n$, problem (\ref{question}) is unconstrained, otherwise constrained. Specifically, when $\Omega$ is a rectangle or hyperrectangle, problem (\ref{question}) is bound (or box) constrained.

When the objective function $f(x)$ is nonconvex, problem (\ref{question}) is often hard to find the global optimum $x^\star$ satisfied
\begin{equation}f(x^{\star})\leq f(x), ~\forall x\in \Omega.\end{equation}
An important reason is that there is no information which can guide to $x^\star$ mathematically. Therefore, many heuristic or evolutionary optimization algorithms were developed for global optimization problems \cite{Floudas-Gounaris-2009,Liu-2013,Weise-2011}.

Through modeling human's brainstorming process, the Brain Storm Optimization (BSO) algorithm proposed recently in \cite{shi2011a, shi2011b} has become a promising population-based evolutionary algorithm, and has attracted more and more theoretical analysis \cite{Cao-Chen-Liu-2017,cheng2016,guo2014,Duan2016,shi2014,shi2015,zhan2013} and practical applications \cite{duan2013,ma2017,duan2014,verma2017,wang2016}. An important progress of BSO is to transform operations in the solution space to the objective space \cite{shi2015}. The new version of BSO is easier in implementation and lower in computational resources on the clustering strategy at each iteration. In this paper, we refer BSO to BSO in the objective space \cite{shi2015}, unless otherwise stated.

In the BSO algorithm \cite{shi2015}, the population evolutes through updating each individual at each iteration. Specifically, the whole population is firstly classified into two categories, the best 20\% individuals are included in the elite population (elites), while the others are included in the set of regular population (normals). Then one or two parent individuals are selected randomly from either or both sets, new child individual is generated through combining these selected parent individuals linearly and then add some white noise. If the new child individual is better than the original one, the original individual will be updated with the new child individual. Such process is then repeated until stopping condition holds.

Many numerical experiments have shown that BSO is good at global exploration but not good enough at local exploitation \cite{Cao-Chen-Liu-2017,Duan2016,Yang2015}. Specifically, BSO can get near optimum quickly, but need much more cost to improve accuracy. Such phenomenon is called the ``degenerated L-curve'' in this paper, and will be discussed in detail next section. The degenerated L-curve phenomenon prevents BSO reaching the optimum quickly with high accuracy, and therefore BSO is not suitable for optimizing directly computational expensive problems or problems needed high accuracy solutions.

The main goal of this paper is to overcome or alleviate the degenerated L-curve phenomenon of BSO. Our strategy is to combine an efficient local solver into BSO. In this way, we hope to accelerate local search firstly and then to improve the whole global search. In this paper, the Nelder-Mead Simplex (NMS) algorithm \cite{NMS1965} is selected, which is one of the most well-known and efficient derivative-free optimization algorithm for local optimization \cite{Ahandani2014,Wang2011,Kao2009}. By suitably fusing the BSO's evolutionary search and simplex search, exploration and exploitation can be well balanced and the whole performance can be improved.

Specifically, BSO is executed for just one iteration (update the population only once), and then the best found position is regarded as a start point for NMS. Given a small budget of computational cost, NMS is executed, and the found best position is then returned back to update the best individual of population. Repeat such process until some stopping condition satisfied. Since BSO is good at global exploration and NMS is good at local exploitation, our strategy can be interpreted as a repeated process including ``finding a good local area'' and ``exploiting the local area''.

The obtained algorithm, Simplex-BSO, is shown to alleviate significantly BSO's degenerated ``L-curve'' phenomenon, and perform significantly better than BSO on some popular benchmark sets. Although one additional parameter is employed, it is shown to be insensitive.

The remainder of this paper is organized as follows. In Section \ref{sec:bso}, the BSO algorithm is reviewed briefly and its degenerated L-curve phenomenon is discussed. Then the Simplex-BSO algorithm is developed and analyzed in section \ref{sec: Simplex-BSO}. Large number of experimental results are reported in Section \ref{sec:ingredients}. Finally, conclusions are drawn in section \ref{sec:conclusion}.

\section{Brain storm optimization and its degenerated ``L-curve''}\label{sec:bso}
In this section, the BSO (Brain storm optimization in the objective space) is reviewed firstly, and then its degenerated L-curve phenomenon is discussed.
\subsection{Brain storm optimization in the objective space}\label{sec:bsoinos}
This version of BSO is proposed in \cite{shi2015}, whose procedure is listed in Algorithm \ref{algo:bso}.

\begin{algorithm}[htbp]
  \caption{\label{algo:bso}Brain Storm Optimization (BSO).}
  Initialization: generate the initial population randomly and evaluate them\;
  \While {stopping conditions do not hold}{
  \textbf{Classification}: classify all solutions into two categories according to their fitness values: the best $20\%$ individuals are called ``elites'' and the others ``normals''\;
  \textbf{Disruption}: select an individual from the population randomly, and change its value in a randomly selected dimension\;

     \textbf{New individual generation}: select one or two individuals from elitists or normal to generate a child\;
     Generate a new child by add a white noise to the child's each dimension\;
     Record the new child if it is better than the current individual\;

  \textbf{Update}: update the whole population;
  }
\end{algorithm}

In Algorithm \ref{algo:bso}, each iteration of BSO includes four components, namely ``Classification '', ``Disruption'', ``New individual generation'' and ``Population updating''. In ``Classification'' stage, the individuals are classified into two clusters according to their fitness values, the best 20$\%$ individuals are clustered as ``elitists'' and the remaining as ``normals''. The component ``Disruption'' is utilized to increase the population diversity, which
is executed usually with a small probability and one individual's value in one random dimension will be replaced by a random number.
The ``Disruption'' strategy is often helpful for individuals to `jump out' of the local optima.

The most important component of BSO is how to generate new individuals. One or two individuals are selected randomly from ``elitists'' or ``normals'' or both, and then the selected individuals are combined linearly to generate a new individual. Then, adding different white noises to the generated new individual in its different dimensions. The author in \cite{shi2015} use the following step-size function (\ref{eq:stepsize})  to control the variances of these white noises,
\begin{equation}\label{eq:stepsize}
\xi(t)={\rm logsig}\big(\frac{0.5\times T-t}{c}\big)\times {\rm rand}(t),
\end{equation}
where logsig() is a logarithmic sigmoid  function, $T$ and $t$ are the maximum and current number of iterations, respectively, $c$ is a coefficient to change logsig() function's slope, and rand() return a random value between $0$ and $1$.

``Population updating'' is utilized to keep good solutions. The new individual generated in the above stage will be recorded if its fitness value is better than the current given individual. However, the current individual is not updated in time until all new individuals are generated and evaluated .In other words, the whole population will be updated at the end of iteration \cite{shi2015}.

These components of the BSO algorithm \cite{shi2015} have been analyzed and improved for specific applications \cite{Cao-Chen-Liu-2017,Abd2017,Peter2017}. In this paper, we aim to improve BSO through overcoming one of its undesirable behaviors.

\subsection{BSO's degenerated ``L-curve''}
It was shown that BSO is promising in solving global optimization problems \cite{Duan2016,shi2011a,shi2011b,zhan2012}. However, BSO is still suffering from balance of global exploration and local exploitation, just like many other global optimization algorithms. On one hand, global exploration provides helpful guidance and is necessary for nonconvex global optimization. On the other hand, local exploitation is very important for searching inside a potential good region and refining the solutions. But how to balance between exploration and exploitation is not easy, and is often algorithm dependent and even problem dependent.

\begin{figure}
\begin{center}
\includegraphics[width=130mm,height=90mm]{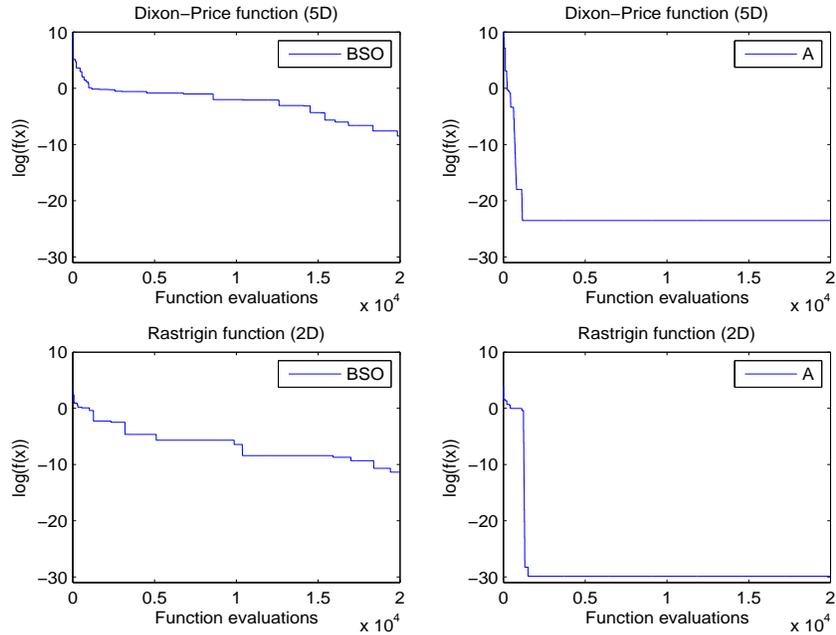}
\caption{Examples of BSO's degenerated L-curve phenomenon. BSO and the algorithm A are adopted to solve the Dixon-Price function and the Rastrigin function, respectively. The found best function values are plotted, and A's curves are ``L-type'' while BSO's are degenerated ``L-type''. }\label{fig-1}
\end{center}
\end{figure}

To judge BSO's ability of balancing between exploration and exploitation, a set of 68 benchmark functions (the Hedar test set \cite{Hedar-2005}, will be presented detailed in Section 4) are tested, and the best function values for each function found by BSO are recorded and plotted. Two examples from the Dixon-Price function and the Rastrigin function are shown in the left subfigures of Figure \ref{fig-1}, where the right subfigures are results from another algorithm (called the algorithm A here for convenience).

Comparing these subfigures, it is clear that algorithm A's curves are L-type with high accuracy (about $10^{-25}$) to the global optimal value 0, which implies that the algorithm A find very good solutions within low computational cost (about 2000 function evaluations). However, BSO's curves decrease slowly and finally (20000 function evaluations) reach solutions with relatively low accuracy (about $10^{-10}$). Therefore, we call it as BSO's degenerated L-curve phenomenon. Among all 68 benchmark functions, there are 56 functions whose curves are similar as those in the left subfigures of Figure \ref{fig-1}. In other words, BSO's degenerated L-curve phenomenon is popular.

Our finding implies that BSO pays less attention to local search, and has degenerated its whole performance. In next section, an efficient local solver is introduced into BSO to overcome or alleviate its degenerated L-curve phenomenon.

\section{Simplex Search Based Brain Storm Optimization}\label{sec: Simplex-BSO}
It is shown that the BSO algorithm proposed in \cite{shi2015} possesses a degenerated L-curve phenomenon, and it is resulted from the weakened local exploitation. In this section, an efficient derivative-free local solver, the Nelder-Mead Simplex method \cite{NMS1965}, is introduced into BSO. Our purpose is to overcome or at least alleviate BSO's degenerated L-curve phenomenon through enhancing its local search.

We firstly review the NMS algorithm briefly, and then develop the Simplex-BSO algorithm.

\subsection{Nelder-Mead Simplex Algorithm}\label{sec:NMS}
The NMS algorithm was proposed by Nelder and Mead in 1965 \cite{NMS1965}, and currently it is still one of the best derivative-free (local) optimization algorithm \cite{Rios-Sahinidis-2013}.

In NMS, a simplex of $\mathbb{R}^n$, which is a geometry with $n+1$ vertical points $x_1, x_2,...,x_{n+1}$, is maintained. At each iteration, the worst point (with worst function value) is often replaced with a new better point, which is generated through reflection, expansion or contraction of the centroid of the best $n$ points around the worst one. If all these operations cannot find a better point, the worst $n$ vertical points shrink around the best one. In this way, the simplex is always updated at each iteration, and the best vertical point will be selected as the solution when some stopping condition holds.

The quasi code of the NMS algorithm is summarized as the Algorithm \ref{algo:NMS}, where the reflection, expansion, contraction (outside or inside) are displayed appear in brackets after the description of the step.

\begin{algorithm}[htbp]
  \caption{\label{algo:NMS} Nelder-Mead Simplex (NMS).}
  Initialization: generate $n+1$ vertices of the initial simplex\;
  \While {stopping conditions do not hold}{
  Order the points from the lowest function value$f(x_1)$ to highest $f(x_{n+1})$\;
  Compute $x_r = 2\bar{x}-x_{n+1}$, where $\bar{x}=\sum_{i=1}^{n}x_i/n$.
\newline \If {$f(x_1)\leq f(x_r)< f(x_n),$} {$x_{n+1}=x_r$, and terminiate this iteration (Reflection)\;}
  \If {$f(x_r)<f(x_1),$} {compute  $x_e = \bar{x}+2 (\bar{x}-x_{n+1})$,
\newline \If {$f(x_e)< f(x_r),$} {$x_{n+1}=x_e$, and terminiate this iteration (Expansion);}
 \Else  {$x_{n+1}=x_r$, and terminate this iteration (Reflection)\;}}
 \If {$f(x_r)\geq f(x_n),$} {
 \If {$f(x_r)<f(x_{n+1}),$} {compute $x_c=(\bar{x}+ x_{r})/2,$
 \newline \If  {$f(x_c)<f(x_r),$}  {$x_{n+1}=x_c$, and terminiate this iteration (Contract outside)}
  \Else  {go to the Shrink step}
  }
\ElseIf{$f(x_r)\geq f(x_{n+1}),$}{compute $x_{cc}=(\bar{x}+ x_{n+1})/2,$
\newline \If  {$f(x_{cc})<f(x_{n+1}),$} {$x_{n+1}=x_{cc}$, and terminiate this iteration (Contract inside).}
 \Else  {go to the Shrink step}}
}
Shrink: $x_i=(x_i+x_1)/2, ~i=2,...,n+1$.
}
\end{algorithm}

At next subsection, the NMS algorithm is introduced into BSO to improve BSO's local exploitation ability.

\subsection{Simplex-BSO: Simplex Search Based Brain Storm Optimization}
To overcome BSO's degenerated L-curve phenomenon, which implies that BSO often gets near the optima quickly but needs much more cost to improve the accuracy, a nature way is to combine an efficient local solver, e.g., the NMS algorithm. The main difficulty of this strategy is how to maximize both BSO's exploration ability and NMS's exploitation ability. After comparing several different designs, we found that the feedback between BSO's global search and NMS's local search is important to improve the whole performance.

Therefore, BSO's exploration is firstly executed, and then NMS's exploitation is run around the found best position. Such process is repeated. Specifically, in each iteration of Simplex-BSO, an iteration of BSO is executed firstly and then the found best position $x_0$ is used as a starting point for NMS's search. The NMS algorithm then begins to search the local area around $x_0$, after consuming $40n$ ($n$ is the dimension of problem) computational cost, the found new best position $x_0^{'}$ is returned back to replace $x_0$. Repeat such process until the stopping conditions hold. The Simplex-BSO algorithm is summarized as the Algorithm \ref{algo:Simplex-BSO}.

\begin{algorithm}[htbp]
  \caption{\label{algo:Simplex-BSO} Simplex-BSO.}
  Initialization: generate the initial population randomly\;
  \While {stopping conditions do not hold}{
  \textbf{Global search}: update the whole population according to the BSO algorithm (Algorithm \ref{algo:bso}). Identify the best individual $x_0$\;
  \While {search cost less than 40*length($x_0$ )}{
  \textbf{Local search}: exploit the search area around $x_0$ through executing the NMS algorithm (Algorithm \ref{algo:NMS}). Let $x_{0}^{'}$ be the found best point\;}
  \textbf{Update}: Update the population via replacing $x_0$  with $x_{0}^{'}$.
  }
\end{algorithm}

As an integration of BSO and NMS, Simplex-BSO provides a better balance between global exploration and local exploitation than BSO or NMS alone. On one hand, BSO's global search helps to find a good starting point for NMS, which has a significant influence on the performance of the NMS method. On the other hand, NMS's efficient local exploitation helps to find desirable solution quickly at the potential good regions.

Then we turn to consider the allocation of the whole computational cost. It is designed to be problem dependent. For each iteration of Simplex-BSO, an iteration of BSO and about $40n$ function evaluations of NMS are executed. In other words, $40n$ plus the size of BSO's population is consumed at Simplex-BSO's each iteration. Since the size of BSO's population is often fixed, therefore, local exploitation is relatively biased as $n$ increases.

\subsection{Influence of the additional parameter}
An additional parameter, $40n$, is introduced in Simplex-BSO to balance BSO's global search and NMS's local search. Although a simple rule is proposed in Algorithm \ref{algo:Simplex-BSO}, it can be set more flexible and even adaptive.

Roughly speaking, low value of the parameter implies that less cost is allocated in local search, and therefore, is more suitable for hard problems or problems with many optima. On the contrary, high value of the parameter is more suitable for relatively easy problems.

However, our extensive experiments shows that the balance parameter is insensitive for most not too hard problems. Therefore, we propose $40n$ in Algorithm \ref{algo:Simplex-BSO} for easy implement. More details about the sensitivity analysis is provided in Section \ref{sec:sensitivity}.

\section{Experimental results}\label{sec:ingredients}
In this section, the Simplex-BSO is compared numerically with the following algorithms:
\begin{itemize}
\item NMS: an algorithm based on Nelder-Mead's simplex search \cite{NMS1965};
\item BSO:  BSO in the objective space \cite{shi2015};
\end{itemize}
Our purpose is to verify that the proposed Simplex-BSO can alleviate significantly BSO's degenerated L-curve phenomenon, and to show Simplex-BSO's good performance.

Two popular sets of benchmark functions, namely the Hedar set \cite{Hedar-2005} and the CEC2017 \cite{CEC2017}, are tested. Some of the hybrid and composition function of the CEC2017 set are really hard to solve, which makes the CEC2017 set harder relatively than the Hedar set.

\subsection{The elimination or alleviation of BSO's degenerated L-curve phenomenon}\label{sec:L-curve-results}
In this subsection, we present the numerical results on the Hedar test set \cite{Hedar-2005}. Table \ref{test problems-1} shows the main information of the Hedar set, including the function names, dimensions, Characteristic, search regions and their minimal function values. There are 16 unimodal problems and 52 multimodal problems in total.

In our experiments, 50 independent runs are executed for each problem, and all algorithms stops only when 20,000 function evaluations are consumed on the Hedar test set.

\begin{table}
\caption{Information about the Hedar test set. }\label{test problems-1}
\begin{center}
\renewcommand\arraystretch{1.2}
\scalebox{0.795}{\begin{tabular}{lllll}
\hline
Function                 &Dimension $n$     &Characteristic     &Search region         &Minimal function value  \\
\hline
Beale                    &2                 &Unimodal           &$[-4.5, 4.5]^2$        &0 \\
Matyas                   &2                 &Unimodal           &$[-8, 12.5]^2$         &0 \\
Sphere                   &2,5,10,20         &Unimodal           &$[-4.1, 6.4]^n$        &0 \\
Sum squares              &2,5,10,20         &Unimodal           &$[-8, 12.5]^n$         &0 \\
Trid                     &6                 &Unimodal           &$[-36, 36]^6$          &-50 \\
Trid                     &10                &Unimodal           &$[-100, 100]^10$       &-200 \\
Zakharov                 &2,5,10,20         &Unimodal           &$[-5, 10]^n$           &0 \\
Ackley                   &2,5,10,20         &Multimodal         &$[-15, 30]^n$          &0 \\
Bohachevsky 1            &2                 &Multimodal         &$[-80, 125]^2$         &0 \\
Bohachevsky 2            &2                 &Multimodal         &$[-80, 125]^2$         &0 \\
Bohachevsky 3            &2                 &Multimodal         &$[-80, 125]^2$         &0 \\
Booth                    &2                 &Multimodal         &$[-100, 100]^2$        &0 \\
Branin                   &2                 &Multimodal         &$[-5,10]*[0,15]$       &0.397887357729739 \\
Colville                 &4                 &Multimodal         &$[-10, 10]^4$          &0 \\
Dixson Price             &2,5,10,20         &Multimodal         &$[-10, 10]^n$          &0 \\
Easom                    &2                 &Multimodal         &$[-100, 100]^2$        &-1 \\
Goldstein and Price      &2                 &Multimodal         &$[-2,2]^2$             &3 \\
Griewank                 &2,5,10,20         &Multimodal         &$[-480, 750]^n$        &0 \\
Hartman 3                &3                 &Multimodal         &$[0, 1]^3$             &-3.86278214782076 \\
Hartman 6                &6                 &Multimodal         &$[0, 1]^6$             &-3.32236801141551 \\
Hump                     &2                 &Multimodal         &$[-5, 5]^2$            &0 \\
Levy                     &2,5,10,20         &Multimodal         &$[-10, 10]^n$          &0 \\
Michalewics              &2                 &Multimodal         &$[0, \pi]^2$           &-1.80130341008983 \\
Michalewics              &5                 &Multimodal         &$[0, \pi]^5$           &-4.687658179 \\
Michalewics              &10                &Multimodal         &$[0, \pi]^{10}$        &-9.66015 \\
Perm                     &4                 &Multimodal         &$[-4, 4]^4$            &0 \\
Powell                   &4,12,24,48        &Multimodal         &$[-4, 5]^n$            &0 \\
Power sum                &4                 &Multimodal         &$[0, 4]^4$             &0 \\
Rastrigin                &2,5,10,20         &Multimodal         &$[-4.1, 6.4]^n$        &0 \\
Rosenbrock               &2,5,10,20         &Multimodal         &$[-5, 10]^n$           &0 \\
Schwefel                 &2,5,10,20         &Multimodal         &$[-500, 500]^n$        &0 \\
Shekel 5                 &4                 &Multimodal         &$[0, 10]^4$            &-10.1531996790582 \\
Shekel 7                 &4                 &Multimodal         &$[0, 10]^4$            &-10.4029405668187 \\
Schkel 10                &4                 &Multimodal         &$[0, 10]^4$            &-10.5364098166920 \\
Shubert                  &2                 &Multimodal         &$[-10, 10]^2$          &-186.730908831024 \\
\hline
\end{tabular}}
\end{center}
\end{table}

\begin{figure}
\begin{center}
\includegraphics[width=130mm,height=90mm]{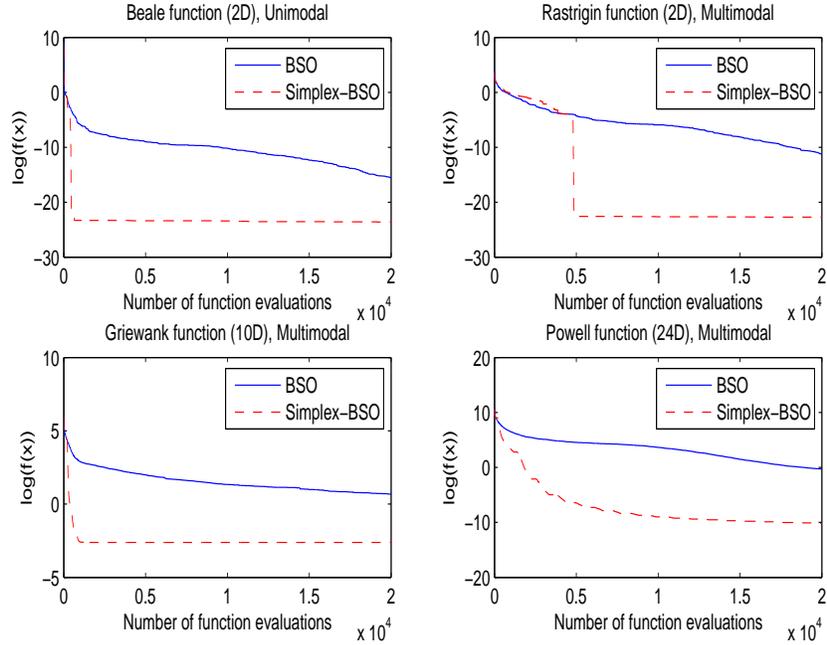}
\caption{L-curves of BSO and Simplex-BSO on 1 unimodal function (Beale) and 3 multimodal functions (Rastrigin, Griewank, and Powell). Simplex-BSO eliminates BSO's degenerated L-curve phenomenon on all 16 unimodal functions, and alleviates it significantly on 40 multimodal functions in the Hedar set.}\label{fig-2}
\end{center}
\end{figure}

To compare the L-curves of BSO and Simplex-BSO, 50 series of the found best function evaluations on each function are averaged and plotted. Fig.\ref{fig-2} shows the L-curves on four representative functions, including 1 unimodal function (Beale) and 3 multimodal functions (Rastrigin, Griewank, Powell).

\subsubsection{L-curves on 16 unimodal functions}
There are 16 unimodal functions in the Hedar set. The L-curve results on these 16 functions are very similar as that of the Beale function (see the top left subfigure in Fig.\ref{fig-2}).

Specifically, Simplex-BSO finds a good solution with accuracy of $10^{-24}$ within less than 1000 function evaluations. On the contrary, BSO achieves an accuracy of $10^{-7}$ within about 2000 function evaluations, and finally achieves the accuracy of $10^{-15}$ when all the 20,000 function evaluations are consumed. Therefore, Simplex-BSO's L-curves are very like ``L'' while BSO's are degenerated.

Since all L-curve results of these 16 functions are very similar, therefore we can conclude that Simplex-BSO eliminates BSO's degenerated L-curve phenomenon on these unimodal functions.

\subsubsection{L-curves on 52 multimodal functions}
There are 52 multimodal functions in the Hedar set. The L-curve results on these 52 functions show that Simplex-BSO still performs very well, and it outperforms BSO on 40 functions. Fig.\ref{fig-2} shows the L-curve results of the Rastrigin (2D), Griewank (10D) and Powell (24D) functions as examples.

Since 20,000 function evaluations are fixed for different dimensional functions, therefore, for low dimensional functions, e.g., the Rastrigin 2D at the top right subfigure, BSO often finds solutions with about $10^{-10}$ accuracy slowly. The L-curve is degenerated. However, Simplex-BSO often can find better solutions with about $10^{-20}$ accuracy and with much faster convergent rate. For higher dimensional functions, e.g., the Griewank 10D and Powell 24D functions, BSO cannot find good solution since the accuracy is larger than $10^0$. On the contrary, Simplex-BSO still can find good solutions with accuracy about $10^{-3}$ for Griewank 10D and $10^{-10}$ for Powell 24D.

However, there still 12 multimodal functions in the Hedar set on which Simplex-BSO is outperformed by BSO. Therefore, we can only conclude that Simplex-BSO alleviates BSO's degenerated L-curve phenomenon on the multimodal functions.

\subsection{Dynamic comparison on the whole set}
Since ``L-curve'' comparison is validation on each function, it is inconvenient and hard to obtain a comprehensive result when the number of benchmark functions is large. Therefore, in this subsection, we provide dynamic comparison results on the whole Hedar set through adopting the visualizing confidence intervals (VCI) method.

\subsubsection{The VCI method}\label{sec:vci}
The VCI method \cite{Liu-Chen-Zhang-2017} is extended from two popular benchmark methods for deterministic optimization algorithms, namely the data profile technique \cite{More-Wild-2009} and the performance profile technique \cite{Dolan-More-2002}. Through visualizing confidence intervals of the found best objective function values, the VCI method is shown to be convenient for benchmarking stochastic global optimization algorithms, especially when the set of benchmark functions or the number of algorithms is large. Hence, it is proposed to replace the traditional statistic test based methods \cite{Liu-Chen-Zhang-2017}.

Specifically, suppose there is a set $\mathbb{S}$ of optimization solvers needed to be compared numerically, and a set $\mathbb{P}$ of benchmark problems is selected. Then given any budget of function evaluations $\mu_f$, run each solver $s\in \mathbb{S}$ on each problem $p\in \mathbb{P}$ for $n_r$ times, and record the series of the found best function values. After all tests finished, a 4-D matrix $H$ with size $\mu_f \times n_r \times n_p \times n_s$ is obtained, where the 4-tuple element $H(k,r,j,i)$ denotes the found best function value during $k$ function evaluations at the $r$-th run when test the $i$-th solver on the $j$-th problem.

The matrix $H$ is then used to generate a sample mean matrix $\overline{H}$
\begin{equation}\label{samplemean}
\overline{H}(k,j,i) = \frac{1}{n_r}\sum_{r=1}^{n_r}H(k,r,j,i)
\end{equation}
and a sample variance matrix $S_H^2$
\begin{equation}
S_H^2(k,j,i) = \frac{1}{n_r-1}\sum_{r=1}^{n_r}\left[H(k,r,j,i)-\overline{H}(k,j,i)\right]^2
\end{equation}
for each algorithm $i=1,...,n_s$ on each problem $j=1,...,n_p$ with $k=1,...,\mu_f$ function evaluations. The confidence upper bound matrix $H_{upper}$ and the confidence lower bound matrix $H_{lower}$ are then defined as follows
\begin{equation}\label{Hupper}
H_{upper}(k,j,i) = \overline{H}(k,j,i)+\frac{2 S_H(k,j,i)}{\sqrt{n_r}},
\end{equation}
\begin{equation}\label{Hlower}
H_{lower}(k,j,i) = \overline{H}(k,j,i)-\frac{2 S_H(k,j,i)}{\sqrt{n_r}}.
\end{equation}

In the VCI method, $\overline{H}, H_{upper}$ and $H_{lower}$ are analyzed statistically with the data profile technique. The data profile is a cumulative distribution function defined for any solver $s\in \mathbb{S}$ as follows
\begin{equation}\label{data_profile}
d_s(\kappa) = \frac{1}{|\mathbb{P}|}\mbox{size}\left\{p\in\mathbb{P}: \frac{t_{p,s}}{D_p+1}\leq \kappa\right\},
\end{equation}
where $|\mathbb{P}|$ denotes the number of test problems, $D_p$ is the dimension of the problem $p$, and size\{\} returns the size of a set. In (\ref{data_profile}), $t_{p,s}$ is the number of function evaluations needed for solver $s$ to find a position $x$ such that the convergence condition
\begin{equation}\label{stoppingcondition}
f(x_0)-f(x)\geq (1-\tau)(f(x_0)-f_L)
\end{equation}
holds, where $x_0$ is the starting point, $\tau>0$ is a tolerance and $f_L$ is the smallest objective function value obtained by any solver within $\mu_f$ of function evaluations. $t_{p.s}=\infty$ if the condition (\ref{stoppingcondition}) does not satisfy after $\mu_f$ function evaluations.

Roughly speaking, the VCI method adopts the data profile technique twice to benchmark stochastic optimization algorithms. Specifically, the data profile technique is used to analyze $\overline{H}$ firstly. From the generated data profiles, a winner solver can be determined in the sense of best average performance. Then the data profile technique is used to compare the winner solver's $H_{upper}$ and the other solvers' $H_{lower}$. The purpose is to confirm that the winner solver performs the best in the sense of the worst deviation.

In summary, the VCI method possesses several advantages than the traditional methods (e.g., statistic test based methods) \cite{Liu-Chen-Zhang-2017}, and is very convenient for our purpose.

\subsubsection{Dynamic comparisons with the VCI method}\label{sec:vci}
The VCI method allows us to compare both stochastic and deterministic optimization algorithms, and therefore is suitable for comparison between BSOs and NMS.
Two steps are needed. Firstly, we compare the average behaviors of BSO, Simplex-BSO and NMS to determine a winner algorithm. Secondly, the winner's $H_{upper}$ is compared with the others' $H_{lower}$ to confirm whether the winner still performs the best at the worst case. If so, then the conclusion is significant statistically that the winner performs better than the other algorithms. Otherwise, the conclusion is that the winner performs averagely better than the other algorithms.

\begin{figure}
\begin{center}
\includegraphics[width=120mm,height=80mm]{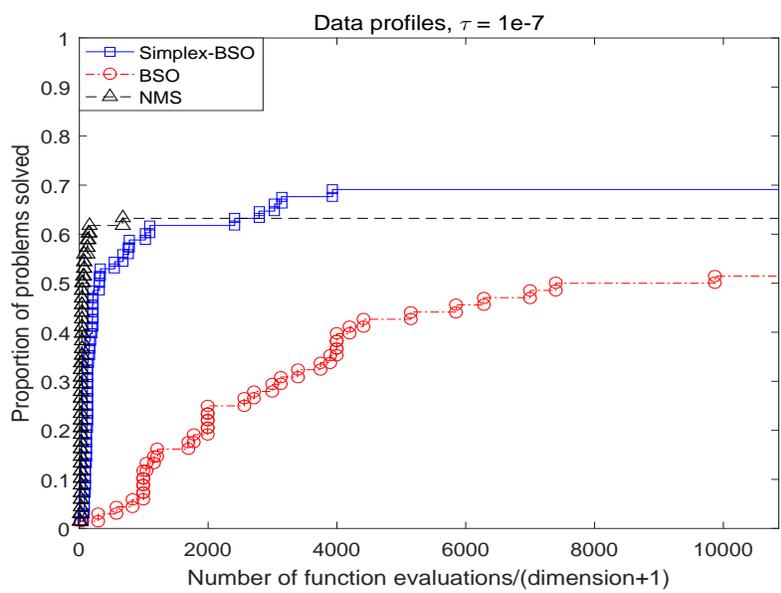}
\caption{Data profiles resulted from comparing the average behaviors of BSO, Simplex-BSO and NMS on the Hedar test set. The results show that Simplex-BSO performs very well since it possesses both NMS's good local search ability and BSO's global search ability.}\label{fig-3}
\end{center}
\end{figure}

Fig.\ref{fig-3} shows the results of the comparison of average behaviors, where the horizontal axis is the computational cost and the vertical axis shows the proportion of problems solved. Therefore, the more close to the left top corner, the better the curve (i.e., the algorithm) is.

We can see from Fig.\ref{fig-3} that the NMS algorithm performs much better than BSO when the computational cost is small, and it can solve more than 60\% problems very quickly. However, NMS cannot perform better when the computational cost is larger than about 2000. On the other hand, BSO performs better and better as the computational cost increases, and finally can solve about 52\% problems. This observation confirms that NMS is an efficient local optimization algorithm while BSO is a global optimization algorithm, and moreover, the original BSO performs much worse than NMS.

From Fig.\ref{fig-3} we can see that Simplex-BSO possesses both NMS's local search ability and BSO's global search ability. When the computational cost is less than about 2000, Simplex-BSO performs worse slightly than NMS. However, Simplex-BSO performs better and better as the computational cost increases. Finally, Simplex-BSO can solve about 69\% problems, about 6\% higher than NMS and 17\% higher than BSO.

The $H_{upper}$ of Simplex-BSO is then compared with $H_{lower}$ of BSO and NMS. The result shows that Simplex-BSO's worst case is outperformed by the best cases of BSO and NMS. According to the VCI method, the conclusion is that Simplex-BSO performs better averagely than both BSO and NMS.

Combined with the results in Section \ref{sec:L-curve-results}, the proposed Simplex-BSO is a promising global optimization algorithm. Through adopting simplex search in BSO, the local search ability is enhanced significantly and the whole global search ability is then strengthened. However, an additional parameter $40n$ is adopted in Simplex-BSO to balance the local search and global search. The sensitivity of the parameter is analyzed in next subsection.

\subsubsection{Sensitivity analysis of the additional parameter}\label{sec:sensitivity}
\begin{figure}
\begin{center}
\includegraphics[width=130mm,height=70mm]{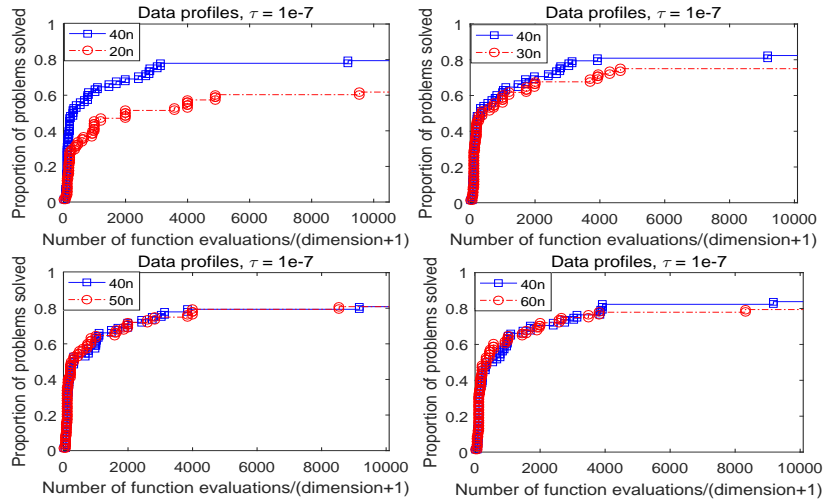}
\caption{Data profiles when comparing different version of Simplex-BSO with different parameter values. The results imply that this parameter is insensitive within a large period $(40n, 60n)$.}\label{fig-4}
\end{center}
\end{figure}

We adopt five different parameter values in Algorithm \ref{algo:Simplex-BSO} to analyze the additional parameter's sensitivity. Specifically, $20n, 30n, 40n, 50n, 60n$ are adopted and tested on the Hedar set. The average behaviors of 50 independent runs are compared with the VCI method, and Fig.\ref{fig-4} shows the results.

From Fig.\ref{fig-4} we can see that Simplex-BSO with $40n$ performs better than the versions with $20n, 30n$, and performs very similar with the versions with $50n, 60n$. Therefore, we conclude that the additional parameter $40n$ is insensitive on the Hedar set, at least in the range of $(40n, 60n)$. There are two reasons why we select the smallest value in this interval. Firstly, all values within $(40n, 50n)$ perform very well. Secondly, a small parameter value is helpful in maintaining population diversity, and therefore enhancing good global search ability.

\subsection{Extended comparison on the CEC2017 test set}
In previous subsection, Simplex-BSO has shown to be able to eliminate BSO's degenerated L-curve phenomenon on unimodal functions and alleviate significantly it on multimodal functions in the Hedar set. Although an additional parameter is introduced, our sensitivity analysis shows that it is insensitive within an large interval.

In this subsection, Simplex-BSO is compared with BSO and NMS on the CEC2017 test set \cite{CEC2017}, which includes 30 different functions and most of them are harder than those in the Hedar set. Our purpose is to verify whether Simplex-BSO still performs well.

Totally, 46 benchmark functions are tested, including all the 16 two-dimensional and 30 ten-dimensional functions in CEC2017 set. Main information about these functions are listed in Table \ref{tab cec17}. For each function, 50 independent runs are executed, and the algorithms stop only when $10,000n$ function evaluations are consumed, where $n$ is the dimension. The VCI method is adopted to analyze the test data.

\begin{table}[htbp]
\caption{Information about the CEC'17 test suite. }\label{tab cec17}
\begin{center}
\renewcommand\arraystretch{1.1}
\scalebox{0.795}{\begin{tabular}{lclc}
\hline
Function                                                &Dimension $n$     &Characteristic     &Minimal function value  \\
\hline
\tabincell{c}{Shifted and Rotated Bent Cigar \\
Function }                                              &2,10               &Unimodal          &100 \\
\tabincell{c}{Shifted and Rotated Sum of \\
Different Power Function }                              &2,10               &Unimodal          &200 \\
\tabincell{c}{Shifted and Rotated Zakharov \\
Function }                                              &2,10               &Unimodal          &300 \\
\tabincell{c}{Shifted and Rotated Rosenbrock¡¯s \\
Function}                                               &2,10               &Multimodal        &400 \\
\tabincell{c}{Shifted and Rotated Rastrigin¡¯s\\
 Function}                                              &2,10               &Multimodal        &500 \\
\tabincell{c}{Shifted and Rotated Expanded \\
 Scaffer¡¯s F6 Function}                                 &2,10               &Multimodal        &600 \\
\tabincell{c}{Shifted and Rotated Lunacek \\
 Bi\_Rastrigin Function}                                &2,10               &Multimodal        &700 \\
\tabincell{l}{Shifted and Rotated Non-\\
Continuous  Rastrigin¡¯s Function}                       &2,10               &Multimodal        &800 \\
\tabincell{c}{Shifted and Rotated Levy \\
Function }                                              &2,10               &Multimodal        &900 \\
\tabincell{c}{Shifted and Rotated Schwefel¡¯s \\
Function }                                              &2,10               &Multimodal        &1000 \\
Hybrid Function 1 (N=3)                                 &10                 &Multimodal        &1100 \\
Hybrid Function 2 (N=3)                                 &10                 &Multimodal        &1200 \\
Hybrid Function 3 (N=3)                                 &10                 &Multimodal        &1300 \\
Hybrid Function 4 (N=4)                                 &10                 &Multimodal        &1400 \\
Hybrid Function 5 (N=4)                                 &10                 &Multimodal        &1500 \\
Hybrid Function 6 (N=4)                                 &10                 &Multimodal        &1600 \\
Hybrid Function 6 (N=5)                                 &10                 &Multimodal        &1700 \\
Hybrid Function 6 (N=5)                                 &10                 &Multimodal        &1800 \\
Hybrid Function 6 (N=5)                                 &10                 &Multimodal        &1900 \\
Hybrid Function 6 (N=6)                                 &10                 &Multimodal        &2000 \\
Composition Function 1 (N=3)                            &10                 &Multimodal        &2100 \\
Composition Function 2 (N=3)                            &10                 &Multimodal        &2200 \\
Composition Function 3 (N=4)                            &2,10               &Multimodal        &2300 \\
Composition Function 4 (N=4)                            &2,10               &Multimodal        &2400 \\
Composition Function 5 (N=5)                            &2,10               &Multimodal        &2500 \\
Composition Function 6 (N=5)                            &2,10               &Multimodal        &2600 \\
Composition Function 7 (N=6)                            &2,10               &Multimodal        &2700 \\
Composition Function 8 (N=6)                            &2,10               &Multimodal        &2800 \\
Composition Function 9 (N=3)                            &10                 &Multimodal        &2900 \\
Composition Function 10 (N=3)                           &10                 &Multimodal        &3000 \\
\hline
\end{tabular}}
\end{center}
\end{table}

\subsubsection{Results on the 2D functions}
\begin{figure}
\begin{center}
\includegraphics[width=110mm,height=73mm]{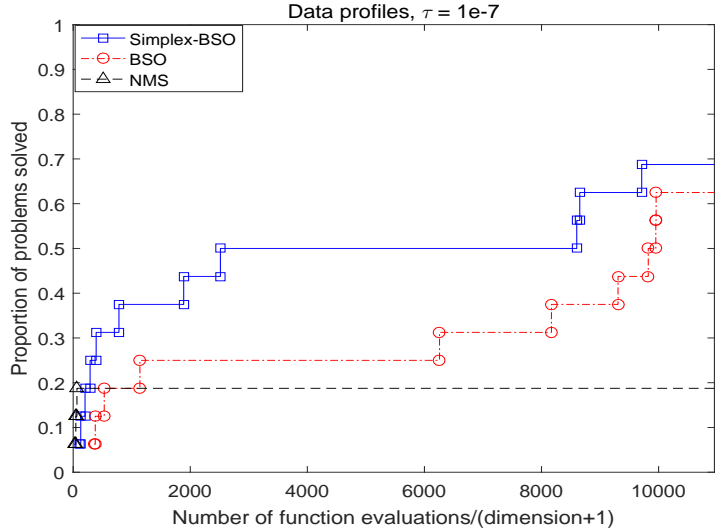}
\caption{Data profile results when comparing Simplex-BSO, BSO and NMS on 16 two-dimensional functions in the CEC2017 set.}\label{fig-5}
\end{center}
\end{figure}

Fig.\ref{fig-5} shows the comparison results on 16 two-dimensional functions in CEC2017. 50 independent runs are averaged, and data profiles are generated through the VCI method.

From Fig.\ref{fig-5} we can see that NMS finds 3 (=16*19\%) functions' optimum very quickly but never gets better. The reason is that NMS is a local solver and often stagnates in a local optimal position. However, both BSO and Simplex-BSO can solve more and more functions as computational cost increases. Finally, Simplex-BSO solves 11 (=16*69\%) functions and BSO solves 10 (=16*63\%) functions, much larger than NMS's 3 functions.

Simplex-BSO performs very well on these functions. On one hand, it solves the most functions among these 3 algorithms. On another hand, it performs the best for any given computational cost. The performance difference between Simplex-BSO and BSO is almost always larger than 6\%, and the difference becomes 25\% when the computational cost between 2500 and 6200.

\subsubsection{Results on the 10D functions}
\begin{figure}
\begin{center}
\includegraphics[width=110mm,height=73mm]{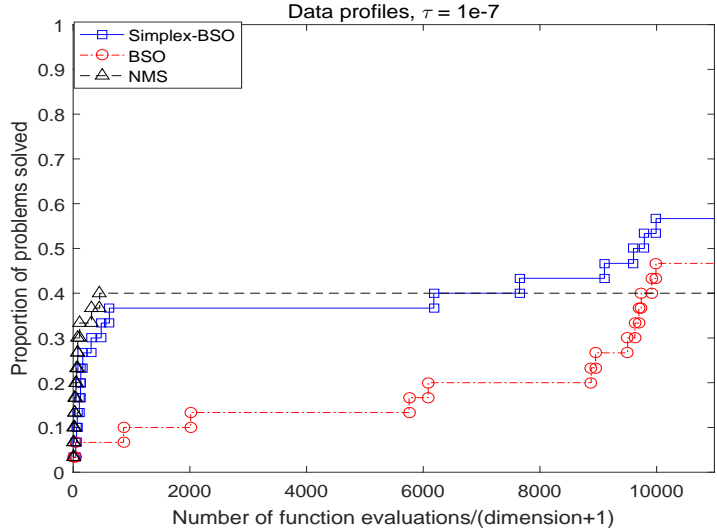}
\caption{Data profile results when comparing Simplex-BSO, BSO and NMS on 30 ten-dimensional functions in the CEC2017 set.}\label{fig-6}
\end{center}
\end{figure}

Fig.\ref{fig-6} shows the comparison results on 30 ten-dimensional functions in CEC2017. 50 independent runs averaged, and data profiles are generated through the VCI method.

From Fig.\ref{fig-6} we can see that NMS solve about 12 (=30*40\%) functions and Simplex-BSO solves 11 (=30*37\%) functions with the best efficiency. On the contrary, BSO only solves 2 functions with the best efficiency.

As the computational cost increases, NMS never solves more functions, while both Simplex-BSO and BSO solves more and more functions. For instance, when the computational cost increases to about 6200, Simplex-BSO solves 12 functions and BSO solves 6 functions. Finally, Simplex-BSO solves 17 (=30*57\%) functions and BSO solves 14 (=30*47\%) functions. Once again, Simplex-BSO performs almost always better than BSO for any given computational cost.

In a word, Simplex-BSO performs still very well averagely on the CEC2017 test set. Combined with the results on the Hedar set, Simplex-BSO is promising global optimization algorithm. Through adopting an efficient NMS local solver in BSO, the obtained algorithm Simplex-BSO enhances its local search significantly and improves its whole performance.

\section{Conclusions}\label{sec:conclusion}
In this paper, the BSO algorithm is shown to possess the degenerated L-curve phenomenon. To overcome this problem, an efficient derivative-free local solver, the Nelder-Mead Simplex method, is adopted into a the BSO algorithm. The obtained algorithm, Simplex-BSO, is shown to be able to eliminate BSO's degenerated L-curve phenomenon on unimodal functions and alleviate this phenomenon significantly on multimodal functions. Although an additional balance parameter is introduced in Simplex-BSO, it is shown to be insensitive within a large interval. Extensive experiments results show that the proposed Simplex-BSO algorithm is a promising global optimization algorithm.

\end{document}